\documentclass[sigconf]{acmart}

\usepackage{hyperref}

\usepackage{listings}
\definecolor{deepblue}{rgb}{0,0,0.5}
\definecolor{deepred}{rgb}{0.6,0,0}
\definecolor{deepgreen}{rgb}{0,0.5,0}
\definecolor{keywords}{RGB}{220,20,60}
\definecolor{strings}{RGB}{0,0,128}
\colorlet{numb}{magenta!60!black}

\lstdefinelanguage{json}{
    basicstyle=\ttfamily\footnotesize,
    commentstyle=\color{strings}, %
    stringstyle=\color{keywords}, %
    showstringspaces=false,
    breaklines=true,
    string=[s]{"}{"},
    comment=[l]{:\ "},
    morecomment=[l]{:"},
    literate=
        *{0}{{{\color{numb}0}}}{1}
         {1}{{{\color{numb}1}}}{1}
         {2}{{{\color{numb}2}}}{1}
         {3}{{{\color{numb}3}}}{1}
         {4}{{{\color{numb}4}}}{1}
         {5}{{{\color{numb}5}}}{1}
         {6}{{{\color{numb}6}}}{1}
         {7}{{{\color{numb}7}}}{1}
         {8}{{{\color{numb}8}}}{1}
         {9}{{{\color{numb}9}}}{1}
}
\usepackage{subcaption}

\usepackage{multirow}
\usepackage[capitalize]{cleveref}

\makeatletter
\def\@ACM@copyright@check@cc{}
\makeatother
\copyrightyear{2025} 
\acmYear{2025} 
\setcopyright{cc}
\setcctype{CC-BY}
\acmConference[SAC '25]{The 40th ACM/SIGAPP Symposium on Applied Computing}{March 31-April 4, 2025}{Catania, Italy}
\acmBooktitle{The 40th ACM/SIGAPP Symposium on Applied Computing (SAC '25), March 31-April 4, 2025, Catania, Italy}\acmDOI{10.1145/3672608.3707732}
\acmISBN{979-8-4007-0629-5/25/03}

\begin{document}
\title{Affordably Fine-tuned LLMs Provide Better Answers to Course-specific MCQs}

\author{Bianca Raimondi}
\orcid{0009-0002-1562-7722}
\affiliation{%
  \institution{University of Bologna}
  \streetaddress{}
  \city{Bologna} 
  \state{} 
  \country{Italy}
  \postcode{}  
}
\email{bianca.raimondi3@unibo.it}

\author{Saverio Giallorenzo}
\orcid{0000-0002-3658-6395}
\affiliation{%
  \institution{University of Bologna
  \\ Bologna, Italy
  \\ INRIA, France}
  \streetaddress{}
  \city{} 
  \state{} 
  \country{}
  \postcode{}  
}
\email{saverio.giallorenzo2@unibo.it}

\author{Maurizio Gabbrielli}
\orcid{0000-0003-0609-8662}
\affiliation{%
  \institution{University of Bologna}
  \streetaddress{}
  \city{Bologna} 
  \state{} 
  \country{Italy}
  \postcode{}  
}
\email{maurizio.gabbrielli@unibo.it}

\renewcommand{\shortauthors}{B. Raimondi et al.}

\begin{abstract}

In education, the capability of generating human-like text of Large Language Models (LLMs) inspired work on how they can increase the efficiency of learning and teaching. We study the affordability of these models for educators and students by investigating how LLMs answer multiple-choice questions (MCQs) with respect to hardware constraints and refinement techniques. We explore this space by using generic pre-trained LLMs (the 7B, 13B, and 70B variants of LLaMA-2) to answer 162 undergraduate-level MCQs from a course on Programming Languages (PL)---the MCQ dataset is a contribution of this work, which we make publicly available. Specifically, we dissect how different factors, such as using readily-available material---(parts of) the course's textbook---for fine-tuning and quantisation (to decrease resource usage) can change the accuracy of the responses. The main takeaway is that smaller textbook-based fine-tuned models outperform generic larger ones (whose pre-training requires conspicuous resources), making the usage of LLMs for answering MCQs resource- and material-wise affordable.
\end{abstract}

\begin{CCSXML}
<ccs2012>
   <concept>
       <concept_id>10010405.10010489.10010491</concept_id>
       <concept_desc>Applied computing~Interactive learning environments</concept_desc>
       <concept_significance>500</concept_significance>
       </concept>
   <concept>
       <concept_id>10010147.10010178.10010179.10010182</concept_id>
       <concept_desc>Computing methodologies~Natural language generation</concept_desc>
       <concept_significance>500</concept_significance>
       </concept>
 </ccs2012>
\end{CCSXML}

\ccsdesc[500]{Applied computing~Interactive learning environments}
\ccsdesc[500]{Computing methodologies~Natural language generation}

\keywords{Education, Large Language Models, Multiple-Choice Questions, Fine-tuning, Quantisation.}

\maketitle

\section{Introduction}
Large Language Models (LLMs)~\cite{zhao2023survey} have recently attracted
attention for their capability to handle complex tasks~\cite{zhang2023survey}.
In education, LLMs can help students and teachers reduce the effort involved in
learning and teaching~\cite{dijkstra2022reading,rasul2023role,extance2023}. In
this paper, we focus on studying how LLMs answer multiple-choice questions
(MCQs) regarding a specific academic topic (bachelor-level programming
languages) with respect to hardware constraints and refinement techniques.

\subsection{Related Work}\label{par:related_work} Recent evidence shows how
ascertaining MCQs is challenging for LLMs and requires ad-hoc prompting to
obtain good results~\cite{RRW23}. Here, we look at work close to ours, focussed
on the domain of question-answering for education. Nori et al.~\cite{NKMCH23}
evaluate GPT-4 on medical competency examinations including both QA and MCQs and
Silva et al.~\cite{SNEAC23} conduct a study on MCQs in the field of agriculture.
In both works, the authors show that using specific prompt engineering
techniques increases the number of correct answers. Other works aim to test LLMs
such as LLaMA and GPT to answer MCQs. In particular, some
works~\cite{trangenerating, savelka2023large, newton_xiromeriti_2023} show that
GPT performs poorly in correctly answering Computer Science MCQs. These results
align with the accuracy profiles we see in ours. Several other
investigations~\cite{pezeshkpour2023large,zheng2023large,zhang2023exploring}
have assessed the ability of LLMs like LLaMA and GPT to handle MCQs. In
particular, \citet{zhang2023exploring} show a higher proficiency of LLMs in STEM
than social sciences and humanities, highlighting the importance of optimising
their knowledge structures. Unfortunately, \citet{zhang2023exploring} conducted
their experiments on three closed-source LLMs, which did not allow them to
refine their models and further study their optimisation.
In order to improve the performance of these models, Bucher and
Martini~\cite{juan2024fine} tried to refine some open source models in the field
of text classification. The authors showed how this type of refinement, carried
out on smaller models, can be used to obtain a resource with a higher precision
than the larger ones. In our contribution, we provide further insight on and
address some limitations of the cited work by 1) experimenting with open-source
LLMs, also comparing different versions thereof (in terms of the number of
parameters) and 2) investigating how one can refine the knowledge structure of
LLMs, via fine-tuning.

\subsection{Proposed Research}
Given the recent, spreading success of LLMs in many assistive tasks, researchers
have been exploring their potential in education. Specifically, LLMs offer
opportunities such as helping students generate ideas, summarise literature, and
create personalised learning experiences. Moreover, they can assist lecturers by
designing quizzes, providing feedback on materials and exams, and generating
educational content tailored to student needs. However, this new technology
comes with challenges, such as the acceptance of AI feedback and the need for
lecturers to integrate these tools into their teaching. For instance, to address
concerns about accuracy and trust, researchers advocate~\cite{extance2023} that
universities coordinate AI-generated and human feedback to ensure reliable
information.

We appraise our work in this context, aiming to find ways to improve the
accuracy of the responses of such models, so that they could be used in real
educational environments, thereby reducing the problem of hallucinations.

Specifically, we investigate the affordability of using LLMs to answer MCQs on a
common item of undergraduate Computer Science curricula: Programming Languages
(PL). We study the accuracy of a publicly available, generic model, LLaMA-2, in
responding 162 MCQs on different subjects of a PL course---we make the dataset
of questions publicly available. In the case of PL, LLMs need to manipulate
linguistic structures that, while different from natural language, are analogous
to the latter and should theoretically be within their capabilities. However,
the domain of PL has technical specificities, similar to other STEM disciplines,
requiring dedicated reasoning skills that the current LLMs may lack. Hence,
broadening our view, we see this work contributing to the general body of
knowledge focused on improving task-specific LLM accuracy. To build our study,
the first question we ask is to what extent one can use the considered, generic
models `as-is' to accomplish the task of ascertaining our MCQ questionnaire. In
answering this question, we dissect how different factors, such as quantisation
(to decrease resource usage), can change the accuracy of the responses.
Following these results, we investigate the affordability of increasing the
accuracy of the model.
Given the work of Bucher and Martini~\cite{juan2024fine}, we wonder whether
fine-tuning could not only improve performance in tasks such as classification
(where models can be tested using classic metrics such as accuracy, precision
and recall), but also be able to acquire more specific knowledge (or at least
refine their handling of the lexicon) in a given subject area. This aspect can
be particularly helpful in the field of Education, where the use of smaller
models, and thus reduced resources, can encourage schools to adopt such
technology. Namely, we refine the model by fine-tuning it with material readily
available to the educator---the textbook used to teach the course---and report
on the hardware requirements and accuracy increases. To further shed light on
how the different parameters of the fine-tuning routine and how the material
from the textbook influences the answers of the refined model, we conduct a
correlation study that clarifies the relationship between these variables and
the increases in accuracy. The LLaMA-2 model comes in four sizes---defined by
their number of parameters; empirically, \textit{ceteris paribus}, the greater
the number of parameters afforded by a model the more accurate are its
responses~\cite{HNADJKPYZ17}. We also explore this dimension by fine-tuning the
two smaller variants 7B and 13B and comparing their accuracy with respect to the
generic largest version 70B. We measure the hardware requirements for refining
7B and 13B, and show how the fine-tuned smaller models achieve equal or greater
accuracy than the largest pre-trained variant.

The paper is structured in the following way: in \cref{sec:methodology}, we
present the dataset, the experimental settings to test the models and the
initial accuracy of the models before fine-tuning, in \cref{sec:results}, we
present the general results and those related to the parameters of fine-tuning,
in \cref{sec:conclusion}, we comment on the results of our work and discuss its limitations.

\section{Methodology}
\label{sec:methodology}
  
\paragraph{MCQ Dataset}
We first present the MCQ dataset.

Both dataset~\footnote{\href {https://zenodo.org/records/14284346}{https://zenodo.org/records/14284346}} and code~\footnote{\href {https://github.com/biancaraimondi/LLaMA2_for_MCQs}{https://github.com/biancaraimondi/LLaMA2\_for\_MCQs}} are publicly available. Data include 162 questions where 15\% are from an undergraduate-level university course on Programming Languages of the degree in Computer Science, 25\% from a book on undergraduate-level questions on PL by Williams et al.~\cite{williams2014mcqs}, and 60\% come from publicly available MCQs from the web (we confirmed the appropriateness of this set of questions with 2 undergraduate PL course teachers).

\begin{figure}[t]
\begin{minipage}{0.5\textwidth}
\begin{lstlisting}[language=json,mathescape]
{ "question": "What problems first inspired the introduction of types?",
"answers": [ "b" ],
"options": { "a": "reference handling",
"b": "set theory paradoxes",
"c": "memory leaks",
"d": "unhandled exceptions" },
"explanation": "The introduction of types was initially inspired [...]",
"topic": "Structuring Data" },
\end{lstlisting}
\end{minipage}
\vspace{-1em}
\caption{MCQs dataset format example (abbreviated \mbox{\color{red}\texttt{"explanation"}} for brevity).\vspace{-1em}}
\label{fig:mcqs_format}
\end{figure}

We format the dataset in JSON, of which we report an excerpt in \cref{fig:mcqs_format}. Technically, the dataset is an array of items that include 5 fields:
\texttt{question}, which contains the text of the MCQ question;
\texttt{options}, which is the list of at least 2 possible answers to the MCQ;
\texttt{answers}, which contains the character identifiers (a, b, c, \dots)
of the correct answers;
\texttt{explanation}, which provides an explanation for the correct answer(s);
\texttt{topic}, which specifies the specific domain of the question with respect to the partition of subjects as found in the chapters of the PL university textbook (we use this datum for fine-tuning and our correlation study).
Since 19 of all 162 questions have multiple correct answer identifiers, the \texttt{answer} field is an array of character identifiers. To check the LLM's answer, it must provide at least one of these correct answers.
\begin{figure}[t]
    \centering
    \includegraphics[width=.5\textwidth]{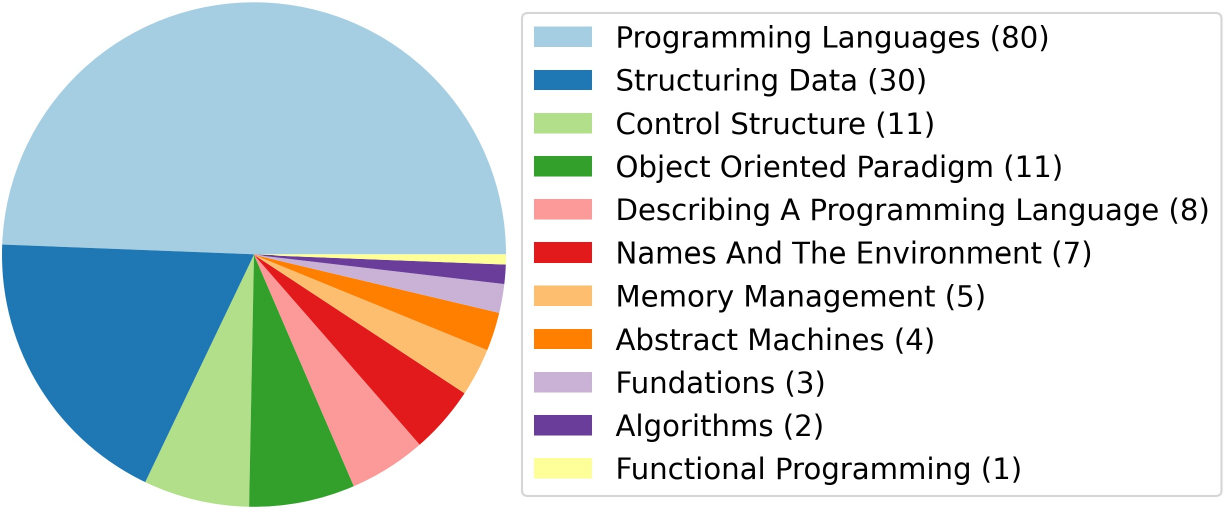}
    \caption{Number of MCQs per partition}
    
    \label{fig:num_MCQs_per_partition}
\end{figure}

We represent the list of subjects with the related number of questions in the MCQ dataset in \cref{fig:num_MCQs_per_partition}.

All subjects except the \textit{Programming Languages} MCQs cover different chapters of the book used for fine-tuning. The latter partition, containing 80 MCQs, covers topics that are still related to the topic of PL but do not appear in the textbook. We include these questions in the dataset as a litmus test to distinguish between questions that a fine-tuned model can answer correctly thanks to the new information provided by the textbook, and questions whose accuracy should not change---but might decrease due to effects like catastrophic forgetting after fine-tuning~\cite{KPRVDRMQRG17}.

\paragraph{Inference with Prompt}
As highlighted in \cref{par:related_work}, previous work showed that LLMs are
less likely to answer correctly questions presented without context, which one
can alleviate by using the technique of prompt engineering (which involves
adding a prompt as a prefix and suffix to the question). Specifically, we use
the pattern

\begin{description}
\item[prefix] - \textit{Answer the following question and provide the correct
answer. The question is a multiple-choice question with multiple correct
answers.}
\item[question] - the question found in the MCQ dataset
\item[options] - the options of the question found in the MCQ dataset 
\item[suffix] - \textit{Answer the question by providing the correct
alternatives. Do not provide an empty answer.}
\end{description}

Given the pattern above, we asked the model variants to generate an answer with
a \emph{temperature} of 0, to minimise the randomness of the
responses.

To calculate the accuracy of a model variant \(v\), we calculate the ratio
\(\frac{a(v)}{q}\) between the correct answers of the tested variant \(a(v)\)
and the total number of questions in the dataset \(q\).

\paragraph{Performance of Generic LLaMA-2 Variants}

We take the 16-bit 7B, 13B, and 70B variants of LLaMA-2~\cite{kalamkar2019study}
and test their accuracy against our dataset---technically, we use them for
inference, i.e., to generate a response to the prompts described above.

We start our analysis of affordability by studying how lighter alternatives of
the variants compare against their base versions. Specifically, we include in
the comparison the quantised alternatives 7B\(_q\), 13B\(_q\), and 70B\(_q\),
which decrease the hardware resources to run inference. To obtain the quantised
alternatives, we use the bits-and-bytes 4-bit NormalFloat quantisation
technique~\cite{dettmers2023qlora}, which substantially decreases the amount of
needed memory while keeping accuracy high (with respect to
same-memory-occupation alternatives like 4-bit Floating Point) and avoid the
more expensive 8-bit alternatives~\cite{dettmers20228bit}

We run our tests on a machine with an Intel Xeon Platinum 8480+ CPU, and an
NVIDIA A100-SXM4, with 80GB of GPU memory---which can host all variants except
the largest model of 70B.

We report the results of our experiments in \cref{tab:answers_percentage}, where
the best-performing variant is 70B\(_q\), with an accuracy of 59\%, followed by
13B, 13B\(_q\), 7B, and 7B\(_q\)---as expected, quantisation reduces the
accuracy of the models, but the loss is reasonable compared to the decreased
memory occupation.

\begin{table}[t]
\centering
\begin{tabular}{|c|c|c|c|c|}
\hline
& \textbf{base} & \textbf{quantised} \\ \hline
\textbf{LLaMA2 7B} & 39\% & 40\% \\ \hline
\textbf{LLaMA2 13B} & 50\% & 45\% \\ \hline
\textbf{LLaMA2 70B} & - & 59\% \\ \hline
\end{tabular}
\caption{Accuracy of the pre-trained LLaMA-2 variants in answering our
dataset.\vspace{-1em}}
\label{tab:answers_percentage}
\end{table}

This result motivates us to look at fine-tuning to increase the accuracy of the different LLaMA-2 variants on the PL domain. We want to test whether giving affordable, readily-available material (i.e., the textbook for an undergraduate-level PL course) we can increase (and to what extent) the accuracy of the variants.

Considering resource consumption, we look at the hardware needed to perform
fine-tuning---this process requires more hardware resources (in
terms of GPU memory) than inference. Specifically, to develop a method
affordable for education using staple data (e.g., the course textbook as a
refinement dataset) and resources obtainable with a small budget (e.g.,
customer-level GPUs, which one can either buy or rent via cloud services). Given
these constraints, we investigate whether we can use the machine used for
inference also to fine-tune the models. In particular, since the 70B model does
not fit into memory for inference, we exclude the usage of that variant for
fine-tuning and rather use the remaining 7B and 13B. 
  
Moreover, since we want to further reduce the required resources, we also look for a way to quantise the fine-tuning process.

\paragraph{On Fine-Tuning: Technique and Dataset}
To fine-tune the selected variants, we use the efficient technique of
Lora~\cite{hu2021lora}---for the quantised alternatives, we use the related
qLora technique~\cite{dettmers2023qlora}. (q)Lora minimises the total number of
parameters updated during fine-tuning thereby saving memory resources.
Concretely, we customise lit-gpt~\cite{litgpt-2023}, a framework that includes
Lora as a fine-tuning technique, to automate the execution of our tests

The dataset we use for fine-tuning draws its contents from the widely adopted undergraduate-level textbook on Programming Language by~\citet{gabbrielli2023programming}, who provided us with private versions of the sources of each chapter, which we adapted for the lit-gpt framework to run fine-tuning.

In particular, we converted the content of the book into a JSON structure
consisting of a sample for each paragraph of the book, made of three different
fields as in the following sample:
\begin{description}
\item[instruction] - \textit{Explain data types}
\item[input] - `` ''
\item[output] - \textit{Data types are present in programming languages for at
least three reasons [...]}
\end{description}
The \textbf{instruction} field indicates the request from the user. We
leave the \textbf{input} field empty (as suggested by lit-gpt's authors)
and we fill the \textbf{output} with the full paragraph text of the book. To
further reduce memory consumption, we split the \textbf{output} field, providing
multiple samples for each paragraph of the book. We split the value of the
\textbf{output} field every 1000 tokens and for each sample generated
we changed the \textbf{instruction} field (e.g. from ``Explain data types'' to
``Explain data types - part 2'').

\paragraph{On Hardware Requirements}
Running our experiments, we tracked the GPU memory usage of the different
LLaMA-2 variants, both considering inference (which requires fewer resources)
and fine-tuning (more resource-consuming). We use this data to frame the
affordability of the different variants. We report the values of
GPU memory occupancy obtained from our experiments in
\cref{tab:resource_consumption}.

\begin{table}[t]
\centering
\begin{tabular}{|c|c|c|c|c|}
\hline
& \multicolumn{2}{|c|}{ \textbf{Inference}} & \multicolumn{2}{|c|}{\textbf{Fine-Tuning}}  \\ \hline & \textbf{base} & \textbf{quant.} & \textbf{base} & \textbf{quant.} \\ \hline
\textbf{7B} & 14GB & 5GB & 45GB & 13GB \\ \hline
\textbf{13B} & 26GB & 10GB & \textcolor{gray}{\textit{>80GB}} & 21GB \\ \hline
\textbf{70B} & \textcolor{gray}{\textit{>80GB}} & 40GB & \textcolor{gray}{\textit{>80GB}} & \textcolor{gray}{\textit{>80GB}} \\ \hline
\end{tabular}
\caption{Memory consumption.\vspace{-1em}}
\label{tab:resource_consumption}
\end{table}

The values in \cref{tab:resource_consumption} allow us to indicate the target
models for this part of the study. Indeed, currently, mid-to-high customer
market video cards sport ca.\@ 24GB. This amount of memory can only fit the
7B\(_q\) and 13B\(_q\) versions for fine-tuning and, hence, we deem only these
two variants affordable for educational institutes (e.g., schools and
universities). Another approach is to split the models across multiple cards or
to use a cloud-based service such as those offered by AWS or Google (assuming
approximately 1\$ per hour). Google Colab, for example, offers a free-of-cost
version with 15GB of GPU memory, which one can use for inference on both 7B and
13B quantised models, and for fine-tuning on the 7B quantised model only. In
this study, we focus on the usage of an on-premises GPU, but we deem the above
alternatives interesting endeavours for future work on usability.
Given our observations on memory occupancy, we consider 7B\(_q\) and 13B\(_q\)
reasonable target models to focus our study on the accuracy and affordability of
using LLMs for answering MCQs. For comprehensiveness, we run our experiments on
all variants we can fit into our machine, reporting and comparing their results.

\paragraph{Experiments Hyperparameters}

To provide a broad overview of the factors that impact the accuracy of LLMs in
answering MCQs, we consider the hyperparameters for inference, reported in
\cref{tab:inference_parameters}, and for fine-tuning, reported in
\cref{tab:finetuning_parameters}. Another crucial parameter involved in our
fine-tuning experiments is the choice of dataset. We deem it important to take
into account how the materials impact the accuracy of the fine-tuned variants.
To this aim, we measure the extent of accuracy improvement derived from
providing topic-specific knowledge to an LLM via fine-tuning. To achieve this
result, we recall that we have two partitioned datasets. One is the dataset of
MCQs, partitioned by the title of the book chapters where they are explained.
The other is the fine-tuning dataset, partitioned into three nested sets on
different content: \emph{a}) the single chapter on ``Structuring Data'', which
corresponds to the largest set of questions related to a specific topic in our
MCQ dataset (cf.~\cref{fig:num_MCQs_per_partition}); \emph{b}) the three
chapters on ``Structuring Data'', ``Abstracting Data'', and ``Object-Oriented
Paradigm'', which include exactly half of the questions covered by the textbook,
and \emph{c}) the whole book (which encompasses a total of 16 chapters). These
partitions allow us to investigate how knowledge found in the different chapters
influences the accuracy of the fine-tuned model. The phenomena we expect to
investigate include the performance of the model with respect to different
amount of information (e.g., if fine-tuning using the whole book decreases
accuracy vs using the chapters regarding specific sets of MCQs) and whether
providing knowledge on neighbouring topics can increase the accuracy of answers
(e.g., fine-tuning with the chapter on ``Structuring Data'' can increase the
accuracy of the ``Object Oriented Paradigm'' MCQs).
In addition, we carefully select hyperparameter values based on insights from
the literature and practical considerations. We test learning rates of 0.001 and
0.0001 to balance convergence speed and stability, with 0.0001 being standard
for LoRA fine-tuning~\cite{parthasarathy2024ultimate}. We range the batch sizes
from 16 to 128 to explore memory usage and gradient stability trade-offs. We
also vary epochs from 2 to 50 to assess both short- and long-term learning
effects, focusing on accuracy and efficiency in resource-limited settings.

\begin{table}[t]
\centering
\begin{tabular}{|c|c|}
\hline
\multicolumn{2}{|c|}{\textbf{Inference}} \\
\hline
\textbf{Model} & [7B, 7B\(_q\), 13B, 13B\(_q\), 70B\(_q\)] \\
\hline
\textbf{Mode} & [pre-trained, fine-tuned] \\
\hline
\end{tabular}
\caption{Inference parameters.}
\label{tab:inference_parameters}
\end{table}

\begin{table}[t]
\centering
\begin{tabular}{|c|c|}
\hline
\multicolumn{2}{|c|}{\textbf{Fine-tuning}} \\
\hline
\textbf{Model} & [7B, 7B\(_q\), 13B\(_q\)] \\
\hline
\textbf{Dataset} & [1 chapt, 3 chapt, whole book] \\
\hline
\textbf{Learning rate} & [0.001, 0.0001] \\
\hline
\textbf{Batch size} & [16, 32, 64, 128] \\
\hline
\textbf{Epochs} & [2, 10, 20, 30, 40, 50] \\
\hline
\end{tabular}
\caption{Fine-tuning parameters.\vspace{-1em}}
\label{tab:finetuning_parameters}
\end{table}

Considering the combinations of hyperparameters, we have 720 fine-tuned variants: 480 of LLaMA-2 7B and 240 of LLaMA-2 13B. 

\paragraph{Treats to Validity}
The main threats to the validity of our results include the representativeness of the MCQs, the material for fine-tuning, and the usage of LLaMA-2. We consider the MCQs in the dataset to be valid because these questions are real MCQs either \emph{a}) used by teachers of a PL course in their exams, \emph{b}) come from a textbook covering the subject or \emph{c}) were collected online but were validated by PL teachers. Of course, we need further studies to be able to generalise our results (e.g., to neighbouring STEM subjects). The material for fine-tuning is valid because it is a widely adopted textbook that educators use to teach PL undergraduate courses, hence authoritative, generally available, and affordable. We deem LLaMA-2 a representative model of current LLMs due to its extensive use in both academia and industry and its performance similarity with respect to current alternatives like Claude-2.1, Gemini Pro, GPT-3.5, Mistral, and Turbo~\cite{JASRMSBetal24}.

\begin{table}[t]
\centering
\begin{tabular}{|c|c|c|c|}
\hline
\multicolumn{4}{|c|}{\textbf{Fine-tuned variants}}\\
\hline
\multicolumn{2}{|c|}{7B} & \multicolumn{2}{|c|}{13B} \\
\hline
\textbf{basic} & \textbf{quantised} & \textbf{basic} & \textbf{quantised} \\
\hline
25\% & 39\% & 21\% & 78\%\\
\hline
\end{tabular}
\caption{Percentage of fine-tuned models that perform better than the pre-trained version on all MCQs.}
\label{tab:general}
\end{table}
\begin{figure}[t]
\centering
\includegraphics[width=.4\textwidth]{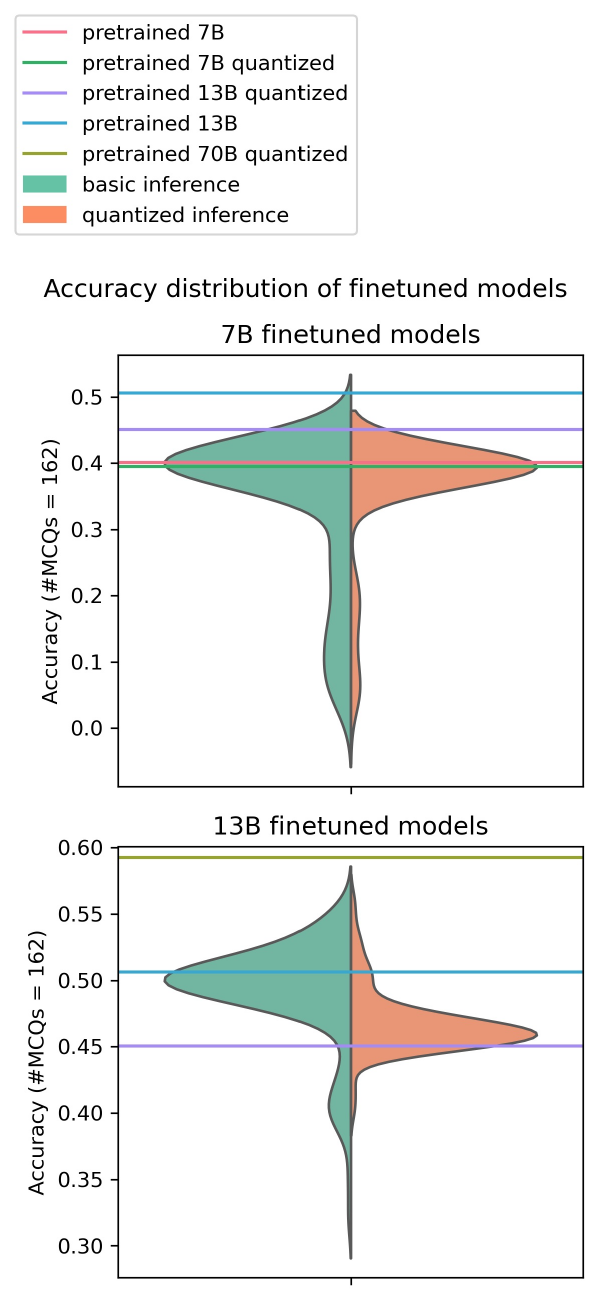}
\caption{Distribution of the accuracy of the fine-tuned models with respect to
the accuracy of the pre-trained models\vspace{-1em}}
\label{fig:general}
\end{figure}

\section{Results and Discussion}
\label{sec:results}
We now present the results of our experiments. Our main goal is to investigate
whether the fine-tuned alternatives have higher accuracy compared to the
pre-trained variants. In \cref{tab:general}, we show the percentage of
fine-tuned models that perform better than the pre-trained version. From the
table, the best-performing fine-tuned variants are those of the 13B quantised
version. While interesting, this datum is quite raw, since we do not know how
these ``better'' fine-tuned variants distribute. Therefore, in
\cref{fig:general}, we report the comparison over all 162 MCQs present in the
dataset to plot the accuracy of these variants against their density.
We illustrate the comparison by graphing the accuracy of the pre-trained
variants as horizontal lines and depicting the distribution of accuracies for
all 720 fine-tuned alternatives using violin plots, which are a hybrid between a
box plot and a kernel density plot, showing summary statistics and the density
of each variable. \cref{fig:general} includes two plots, resp.\@ for the 7B and
13B variants. In the figure, the y-axis represents the accuracy of the
fine-tuned models, while the width of the violin at different y-axis values
indicates the density of the data points, i.e., the distribution of the
accuracies. This visual representation highlights where data points are
concentrated, providing insight into the distribution's shape, which is crucial
for understanding the fine-tuning behavior in this study. In the figure, we also
show how quantisation affects accuracy, reporting on the left of the plots the
base fine-tuned variants and on the right quantised variants. For reference, we
compare both sides with their pre-trained versions (the pink horizontal line for
7B, and the blue horizontal line for 13B). 

Looking at the plots, a first observation is that there is a part of
fine-tuned models that have greater accuracy than the one of the pre-trained
variant. This happens both for 7B and 13B models. Given this result, we also
compare the accuracy of the fine-tuned alternatives of a given variant with the
accuracy of the immediately larger pre-trained version. Namely, we draw the
accuracy line of the pre-trained 13B and 13B\(_q\) in the plot of the fine-tuned
7B alternatives and the accuracy line of the pre-trained 70B\(_q\) variant (the
green horizontal line) in the plot of the fine-tuned 13B alternatives. This
comparison allows us to spot that some fine-tuned alternatives can achieve the
same accuracy as the larger/base versions. 

Looking at quantisation, from \cref{fig:general} we can notice that, as
expected, the quantised models (both pre-trained and fine-tuned) tend to be less
accurate, mostly for the 13B version/variants.

Another important consideration concerns catastrophic forgetting, which, from
\cref{fig:general}, seems to affect about 15\% (70 of 480 models) of the
fine-tuned 7B alternatives suffer from the problem. The observed phenomenon
aligns with reports in the literature~\cite{KPRVDRMQRG17} where smaller
variants, which have a low accuracy compared to the bigger pre-trained
alternatives, tend to forget how to deal with concepts seen during the
pre-training phase. To break the issue down, below, we first show how different
variants and alternatives answer different domain questions---thanks to the
partitions of the MCQs dataset. Then, we study how hyperparameters influence the
results, concluding with a correlation study between the hyperparameters and the
accuracy of the partitions.

\subsection{Topic-wise Accuracy}

\begin{figure*}[t]
    \begin{center}
        \begin{minipage}[t]{.3\textwidth}
            \centering
            \includegraphics[width=\textwidth]{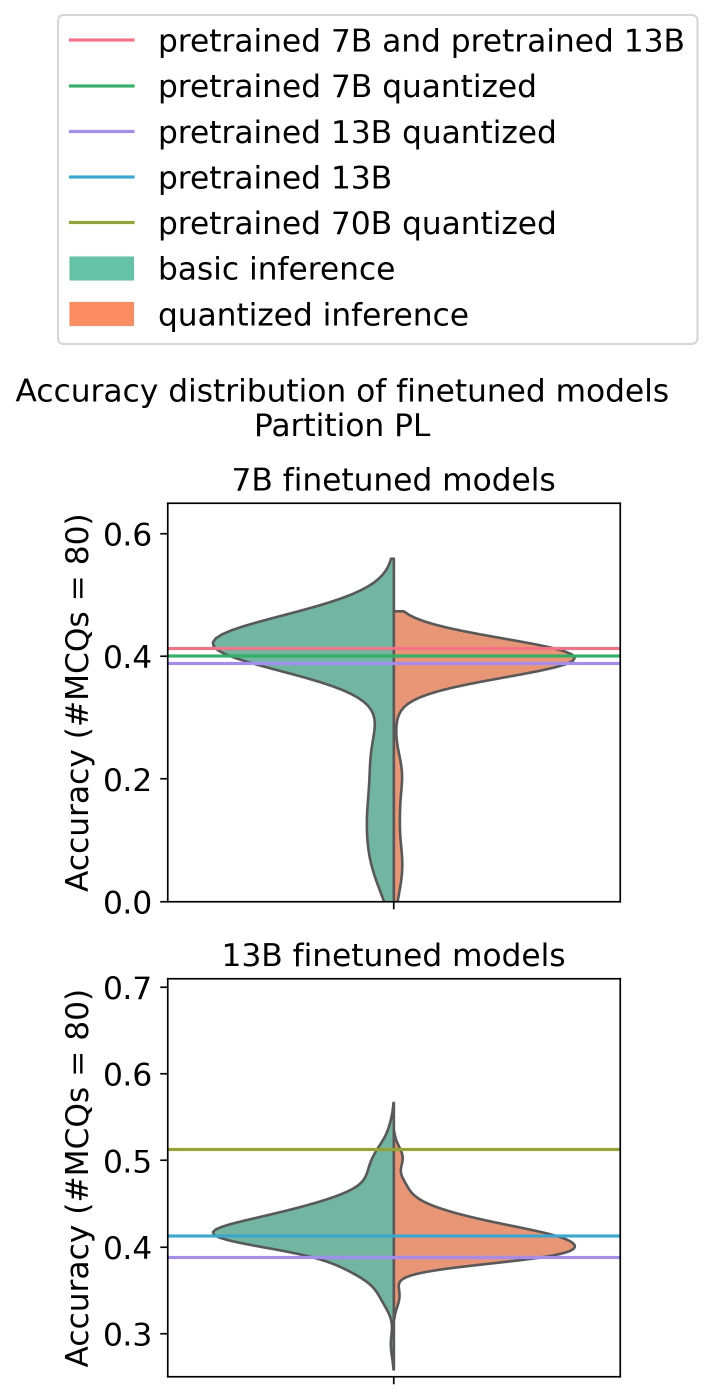}
            \caption{\textbf{PL} MCQs Acc.}
            \label{fig:comp_pl}
        \end{minipage}
        \begin{minipage}[t]{.288\textwidth}
            \centering
            \includegraphics[width=\textwidth]{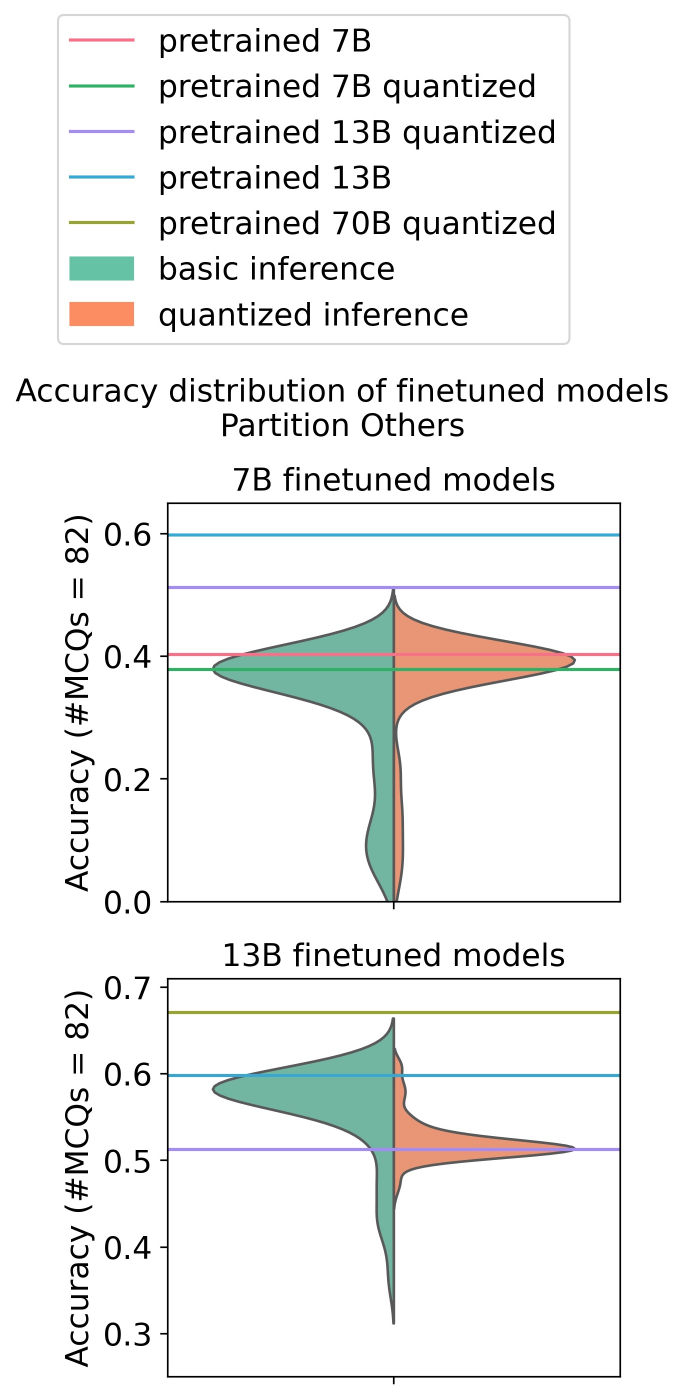}
            \caption{\textbf{Others} MCQs Acc.}
            \label{fig:comp_others}
        \end{minipage}
        \begin{minipage}[t]{.36\textwidth}
            \centering
            \includegraphics[width=\textwidth]{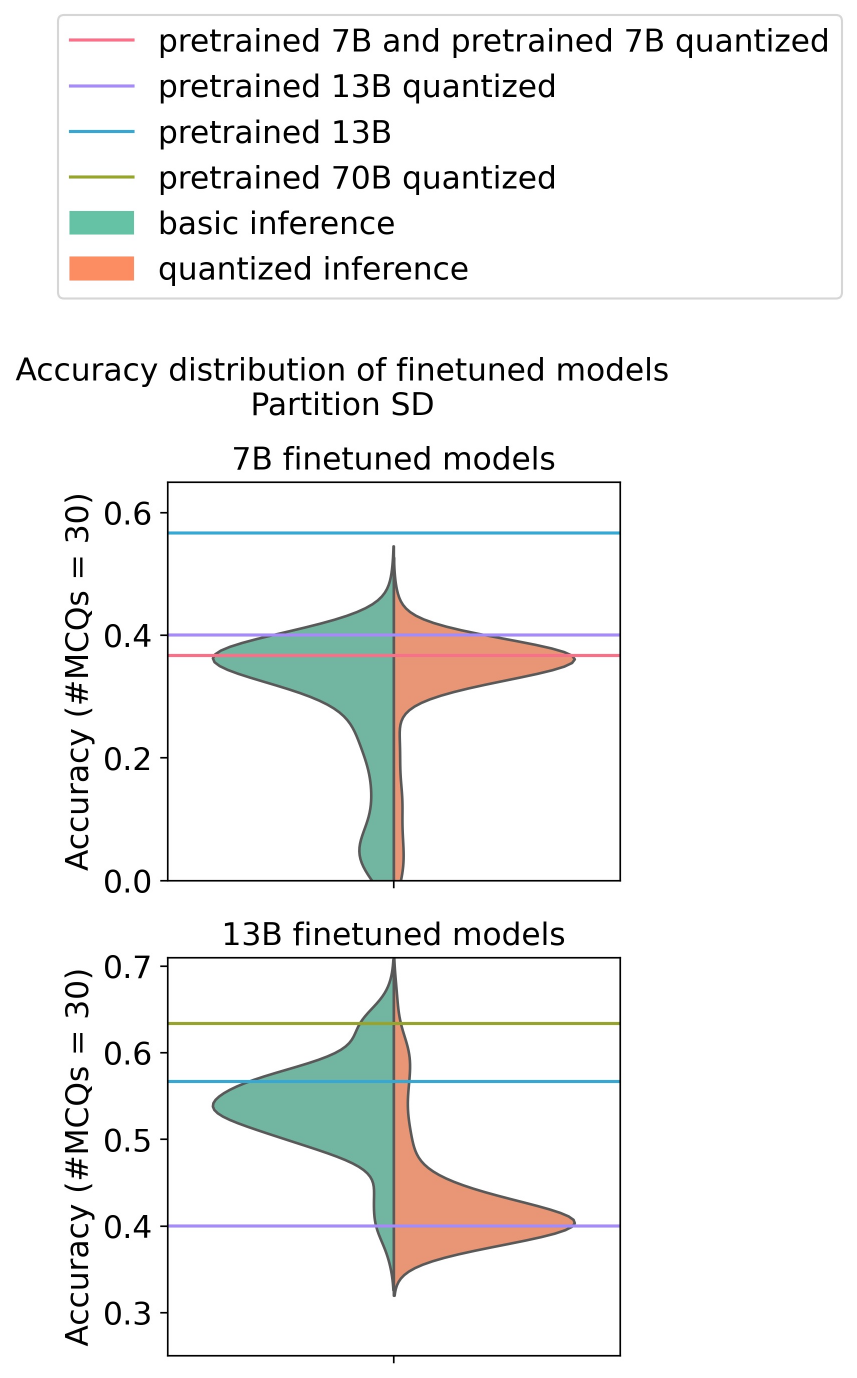}           \caption{\textbf{SD} MCQs Acc.}
            \label{fig:comp_sd}
        \end{minipage}
        \label{fig:PL_vs_others}
    \end{center}
    \vspace{-1em}
\end{figure*}

We start by looking at three partitions of the MCQs, namely, the set of
questions not covered by the book used for fine-tuning (\textit{PL}), the dual
set of questions covered by the book (\textit{Others}), and the largest set of
questions covered by the book included in \textit{Others} (\textit{SD} -
``Structuring Data''). We resp.\@ report the accuracy performance of all the 720
fine-tuned alternatives in \cref{fig:comp_pl,fig:comp_others,fig:comp_sd}, using
the same visualisation template of \cref{fig:general}. First, we notice
that 
the accuracy of the pre-trained versions is higher for \textit{Others}
(\cref{fig:comp_others}) than for \textit{PL} (\cref{fig:comp_pl}) MCQs. The
results show that for the \textit{PL} partition, the fine-tuned models can
achieve higher accuracy than the pre-trained variants, both for 7B and 13B. The
increase for \textit{Others} is smaller since the pre-trained variants start
from a higher accuracy with respect to \textit{PL}. We can notice, e.g., that
the pre-trained accuracy of the 13B version in \textit{PL} and \textit{Others}
differs of ca. 15\%. The explanation of the increase in accuracy for \textit{PL}
MCQs---whose content is not covered by the material used for fine-tuning---is
that the fine-tuned alternatives ``learn'' to deal with the vocabulary of these
MCQs since they share the domain of the content for fine-tuning (on Programming
Languages). 

With the \textit{SD} partition (\cref{fig:comp_sd}), we investigate a subset of
the \textit{Others} partition. The results show that, considering only
\textit{SD} questions, we have a higher accuracy for the fine-tuned
alternatives---reaching ca.\@ 70\%. The comparison between \textit{Others} and
\textit{SD} allows us to infer that some characteristics of the questions or the
related content in the fine-tuning material lent itself better than the whole
set of \textit{Others} MCQs to fine-tuning. We start this analysis by looking
into the influence of the hyperparameters on the accuracy of the \textit{SD}
partition.

\subsection{Hyperparameters of the \textit{SD} partition} \label{sec:results_SD_hyperparameters}

We now investigate which parameters influence the models' accuracy, identifying which factors enable fine-tuned variants to achieve better accuracy than their pre-trained base versions.

\subsubsection{Fine-tuning dataset}
The first hyperparameter we investigate is the influence of the fine-tuning dataset on accuracy. In general, we want to measure how much knowledge from the book the fine-tuned variants need to ``see'' to increase their accuracy.

As mentioned, the fine-tuning process we followed using three sets of chapters
from the book supports this type of analysis. Recalling the partitioning, the
first set of chapters includes only the one directly related to the \textit{SD}
MCQs partition, i.e., ``Structuring Data'', the second set includes
``Structuring Data'' plus the neighbouring chapters of  ``Abstracting Data'' and
``Object-Oriented Paradigm'', and the third set includes all chapters of the
book.

\begin{table}[t]
\centering
\begin{tabular}{|c|c|c|c|c|}
\hline
& \multicolumn{2}{|c|}{\textbf{7B}} & \multicolumn{2}{|c|}{\textbf{13B}} \\
\hline
& \textbf{basic} & \textbf{quant.} & \textbf{basic} & \textbf{quant.} \\
\hline
\textbf{1 chapter} & 4\% & 4\% & 20\% & 43\% \\
\hline
\textbf{3 chapters} & 4\% & 5\% & 13\% & 43\% \\
\hline
\textbf{whole book} & 3\% & 5\% & 7\% & 28\% \\
\hline
\end{tabular}
\caption{Percentage of fine-tuned models that outperform the pre-trained version on \textbf{SD} MCQs with models fine-tuned with different book chapters.\vspace{-1em}}
\label{tab:finetuning_dataset}
\end{table}

We report in \cref{tab:finetuning_dataset} the percentage of fine-tuned variants
that outperform their pre-trained version on the \textbf{SD} MCQs, divided
according to the partitions of the book used for their fine-tuning. From the
results, only one chapter of the book can generally lead to more stable results
(i.e., the accuracy of all alternatives falls within a narrower range than using
the other two sets) without losing higher accuracy results. This is particularly
evident for the 13B variants, which have a higher number of fine-tuned
alternatives that reach an accuracy that is higher than the one of pre-trained
models.

\subsubsection{Quantised fine-tuning}
\label{sec:quantised_fine_tuning}
\begin{table}[t]
\centering
\begin{tabular}{|c|c|c|}%
\hline
& \multicolumn{2}{|c|}{\textbf{7B}} %
\\
\hline
& \textbf{basic} & \textbf{quant.} %
\\
\hline
\textbf{basic fine-tuning} & 0\% & 4\% %
\\
\hline
\textbf{quantised fine-tuning} & 7\% & 5\% %
\\
\hline
\end{tabular}
\caption{Percentage of fine-tuned models that outperform the pre-trained version on SD MCQs with models fine-tuned with or without quantisation.\vspace{-1em}}
\label{tab:finetuning_quantized}
\end{table}
Since we test variants fine-tuned with or without quantisation, we explore this
dimension by reporting in \cref{tab:finetuning_quantized} the results. In the
table, the 7B  fine-tuned quantised variants are more likely to improve the
accuracy of the results. We can only make this observation for the 7B variants
since our machine could only host in memory the latter in both base and
quantised forms, while we could fine-tune only the quantised version of 13B.
This result is important for our purpose, since it shows that one can fine-tune
the models with a smaller memory requirement, thus making it more affordable.

\subsubsection{Learning Rate}
\begin{table}[t]
\centering
\begin{tabular}{|c|c|c|c|c|}
\hline
& \multicolumn{2}{|c|}{\textbf{7B}} & \multicolumn{2}{|c|}{\textbf{13B}} \\
\hline
& \textbf{basic} & \textbf{quant.} & \textbf{basic} & \textbf{quant.} \\
\hline
\textbf{0.001} & 0\% & 9\% & 17\% & 55\% \\
\hline
\textbf{0.0001} & 7\% & 0\% & 10\% & 20\% \\
\hline
\end{tabular}
\caption{Percentage of fine-tuned models that outperform the pre-trained version on SD MCQs with models fine-tuned with different learning rates.\vspace{-1em}}
\label{tab:learning_rate}
\end{table}
Similarly, we explore the influence of learning rate on fine-tuning, divided
between 0.001 and 0.0001. The results, in \cref{tab:learning_rate}, show
that for the 7B variants a higher learning rate favours quantisation while a
lower learning rate favours the base variants. 
Moreover, the 7B variants show catastrophic forgetting and a learning rate
inversely proportional to accuracy, while the 13B variants improve in direct
proportion to the learning rate.

\subsubsection{Batch Size \& Epochs}
\begin{figure}[t]
    \centering
    \includegraphics[width=0.5\textwidth]{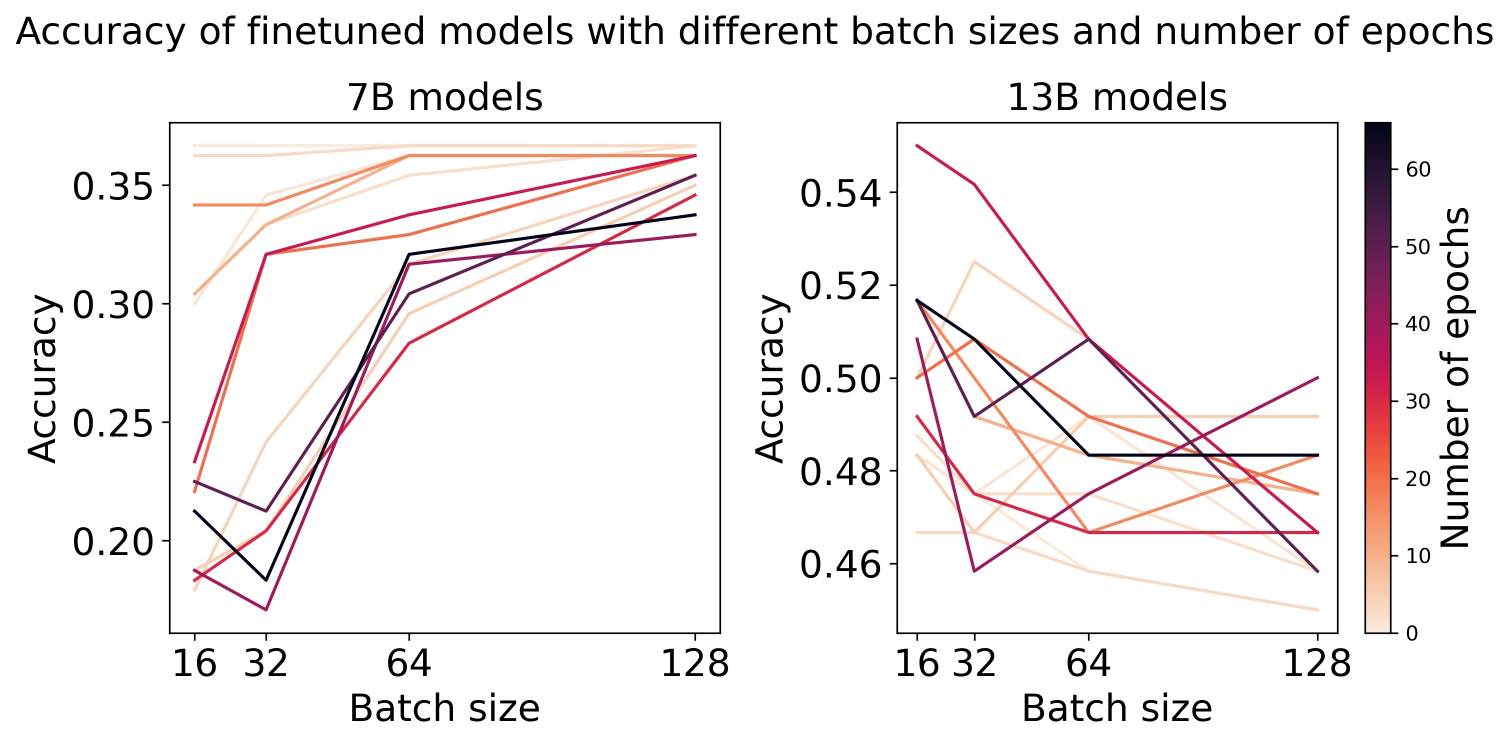}
    \caption{Batch sizes accuracy with respect to epochs.\vspace{-1em}}
    \label{fig:epochs_batchsize}
\end{figure}
We investigate the influence on fine-tuning of batch sizes and numbers of epochs
together. We plot our results in \cref{fig:epochs_batchsize} dividing between
the 7B and the 13B fine-tuned variants. In the plots, we show the relationship
between batch size and accuracy depending on the number of epochs. What emerges
from the plots is that we have a dual behaviour for the two versions of the
models. The 7B variants achieve high accuracy with high batch sizes and a low
number of epochs. Contrarily, the 13B variants achieve high accuracy with
smaller batch sizes and a higher number of epochs. The latter result shows that
the 7B variants, given the lower number of parameters, are less prone to
increase in accuracy with a higher number of iterations, probably due to their
reduced parameter space. This observation leads us to notice that the
fine-tuning process is more beneficial for larger models that afford a larger
parameter space to store new, domain-specific knowledge.

\subsection{Correlations}
\begin{figure}[t]
    \centering
    \includegraphics[width=0.5\textwidth]{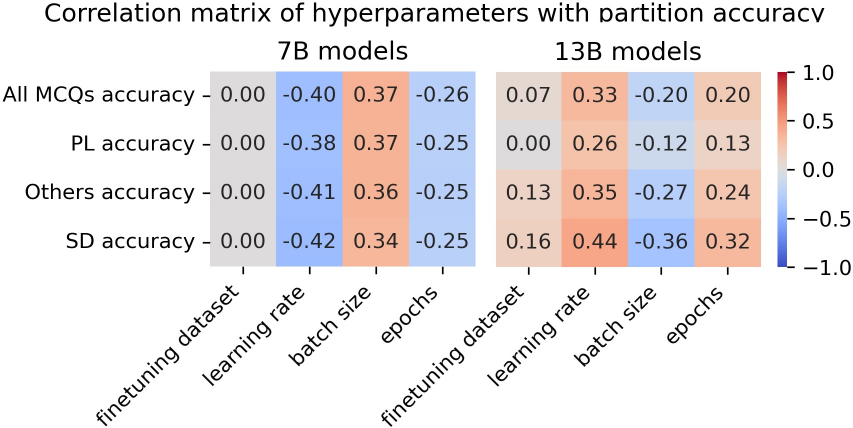}
    \caption{Correlation Matrix of the hyperparameters with accuracy of different MCQs dataset partitions.\vspace{-1em}}
    \label{fig:correlation_matrix}
\end{figure}
The last part of our experiments regards studying the correlation between the hyperparameters and the accuracy of the MCQ dataset partitions, reported in
\cref{fig:correlation_matrix}. The results confirm the dual behaviour observed
earlier: hyperparameters that positively correlate with the 7B variants
negatively correlate with the larger one and vice versa. In particular, notice
the column of the fine-tuning dataset. We obtained this column by labelling the
fine-tuning dataset hyperparameter to range from 1 to 3, where 1 identifies the
one spanning the entire book, 2 three chapters, and 3 the single chapter on
\textbf{SD} (i.e., the greater the number the more specific the information).
From \cref{fig:correlation_matrix}, we observe that this hyperparameter is neutral for the 7B variants but positively correlated for the 13B variants, indicating that fine-tuning with specific MCQ information significantly increases model accuracy (e.g., 0.16 for the \textit{SD} partition).

\section{Discussion and Conclusion}
\label{sec:conclusion}
We investigated the affordability of using LLMs in education, focusing on
LLaMA-2, MCQs, and the domain of PL. Our most general result is that one can
obtain good accuracy in answering domain-specific MCQs with small fine-tuned
models (7B) that run on consumer-grade machines, making it affordable for
education purposes. These results are particularly interesting because the
smaller fine-tuned models achieve an accuracy that is similar to the larger
pre-trained ones. More in detail, as observed in
\cref{sec:results_SD_hyperparameters}, we characterised how fine-tuning
hyperparameters influence accuracy, finding that the choice of the dataset (in
our case, which chapters of the textbook) has the highest influence, along with
quantisation performed during the fine-tuning. Our study has confirmed that some
smaller fine-tuned LLMs are prone to catastrophic forgetting. To address this
issue, we conducted an extensive analysis of the hyperparameters that affect
accuracy during fine-tuning. These results help future research in avoiding
repeating less adequate approaches and focus on more efficient fine-tuning
methods. Another interesting perspective regarding education relates to
understanding how LLMs generate responses, offering valuable insights into their
decision-making processes and potentially improving their accuracy and
reliability as tutoring systems.

\subsection{Limitations and Future Work}

\paragraph{Scope and Experiments.}
In this study, we focussed on the subject of Programming Languages, which limits
the coverage of our results. To be able to extend our results to other areas
(and generalise them), we would need to perform the same study on both
neighbouring subjects (e.g., from mathematics, physics, biology, and
engineering) but also more distant ones (e.g., from literature, history,
philosophy, and arts). Broadening our focus on tutoring systems, an interesting
evolution of this work regards the usage of multimodal models---e.g., such that
MCQs and fine-tuning datasets can integrate visual elements---to both cover more
MCQs types and enhance the accuracy of LLMs using textbooks, which frequently
use figures and schemas to integrate explanations.

\paragraph{Datasets and Models.}
While our study sheds light on the affordability and effectiveness of using LLMs
in educational contexts, there are notable limitations worth acknowledging.
Firstly, regarding the fine-tuning dataset, our research primarily focused on
specific chapters of the textbook, which may limit the generalisability of our
findings to broader educational content that falls outside the information
covered in the book. The selection of these chapters can influence the
performance of the fine-tuned models. Exploring a more diverse array of
educational materials can provide a more comprehensive understanding of LLMs'
efficacy across various subjects and topics. Additionally, our investigation
into multiple-choice questions (MCQs) datasets is limited in scope. Further
research can delve deeper into the nuances of MCQs from different disciplines,
considering factors such as question complexity and diversity. Addressing these
limitations can enhance the applicability and robustness of our findings in
real-world educational settings. Regarding the models used, the limitations
concern the availability of the latter as open-source resources for use and
testing (e.g., GPT is not open-source).

\paragraph{Affordability for Institutions}
Looking at affordability for educational institutions with limited resources, it is important to acknowledge that, while fine-tuning smaller models like the 7B and 13B variants is feasible on mid-range GPUs, larger models are currently impractical due to their significant/costly hardware demands. However, resource-saving strategies, such as LoRA and parameter-efficient fine-tuning (PEFT)~\cite{peft}, can mitigate these challenges by reducing memory and computational requirements. Additionally, institutions can explore cloud-based GPU solutions, including AWS and Google Colab, which offer scalable access to high-performance hardware at competitive rates. These alternatives broaden access to fine-tuning capabilities, making advanced model training more affordable and inclusive for academic purposes.

\begin{acks}
We acknowledge CINECA\footnote{\href {https://www.cineca.it}{https://www.cineca.it}.} for the availability of computing resources and support.
  
This work was supported by Future AI Research (FAIR) PE01, SPOKE 8 on PERVASIVE AI funded by the National Recovery and Resilience Plan (NRRP).
\end{acks}

\bibliographystyle{ACM-Reference-Format}
\bibliography{biblio} 

\end{document}